\pdfoutput=1

\documentclass[11pt]{article}

\usepackage[final]{coling}

\usepackage{times}
\usepackage{listings}
\usepackage{latexsym}
\usepackage{amsmath}
\usepackage{algorithm}
\usepackage{algorithmic}
\usepackage{subfigure}
\usepackage{subcaption}
\usepackage[T1]{fontenc}

\usepackage[utf8]{inputenc}

\usepackage{microtype}

\usepackage{inconsolata}

\usepackage{graphicx}

\usepackage{multirow}

\title{Reasoning Graph Enhanced Exemplars Retrieval for In-Context Learning}

\author{Yukang Lin\textsuperscript{\rm 1}, \hspace{0cm}
        Bingchen Zhong\textsuperscript{\rm 1}, \hspace{0cm}
        Shuoran Jiang\textsuperscript{\rm 1}, \hspace{0cm}
        Joanna Siebert\textsuperscript{\rm 1}, \hspace{0cm}
        Qingcai Chen\textsuperscript{\rm 1,2}\footnotemark[2] \\
        \textsuperscript{\rm 1}Harbin Institute of Technology (Shenzhen), Shenzhen, China\\
        \textsuperscript{\rm 2}Peng Cheng Laboratory, Shenzhen, China \\
        \texttt{\{yukang.lin00,shuoran.chiang\}@gmail.com} \  
        \texttt{\{zhongbingchen\}@stu.hit.edu.cn} \\  
        \texttt{\{joannasiebert\}@yahoo.com} \  
        \texttt{\{qingcai.chen\}@hit.edu.cn}
}

\begin{document}
\maketitle

\renewcommand{\thefootnote}{\fnsymbol{footnote}} 
\footnotetext[2]{Corresponding author.} 
\renewcommand{\thefootnote}{\arabic{footnote}}

\begin{abstract}
Large language models (LLMs) have exhibited remarkable few-shot learning capabilities and unified the paradigm of NLP tasks through the in-context learning (ICL) technique. Despite the success of ICL, the quality of the exemplar demonstrations can significantly influence the LLM's performance. Existing exemplar selection methods mainly focus on the semantic similarity between queries and candidate exemplars. On the other hand, the logical connections between reasoning steps can also be beneficial to depict the problem-solving process. This paper proposes a novel method named Reasoning Graph-enhanced Exemplar Retrieval (RGER). RGER first queries LLM to generate an initial response and then expresses intermediate problem-solving steps to a graph structure. After that, it employs a graph kernel to select exemplars with semantic and structural similarity. Extensive experiments demonstrate the structural relationship is helpful to the alignment of queries and candidate exemplars. The efficacy of RGER on mathematics and logical reasoning tasks showcases its superiority over state-of-the-art retrieval-based approaches.  Our code is released at \href{https://github.com/Yukang-Lin/RGER}{https://github.com/Yukang-Lin/RGER}. 
\end{abstract}

\section{Introduction}
With the increasing scale of large language models (LLMs), in-context learning (ICL) has emerged as a striking property \cite{gpt3}. ICL enables language models to perform unseen tasks through prompts consisting of a few demonstrations, without requiring any gradient updates \cite{iclsurvey}. Despite this notable capability, the effectiveness of ICL can be highly sensitive to the selection of templates, verbalizers, and demonstrations \cite{iclchange}. In addition, the quality of selected demonstrations can significantly impact the ICL's performance, which makes in-context exemplar selection a fundamental problem \cite{liu2021, wu2022self}.


\begin{figure*}[!htbp]
  \centering
  \includegraphics[width=\linewidth]{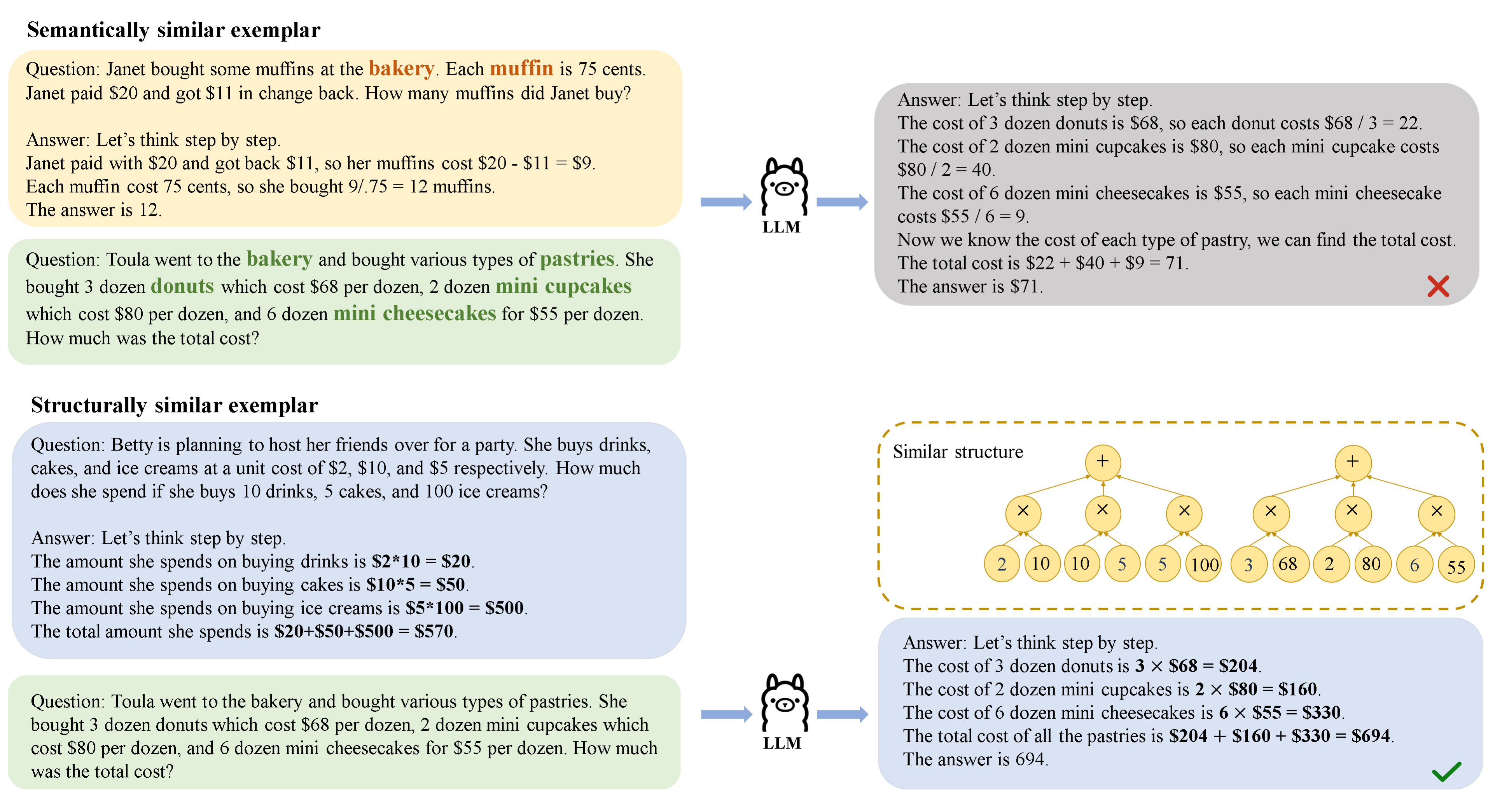}
  \caption {Comparison between semantically similar(top) and structurally similar(bottom) for in-context demonstrations. (Top)The exemplar with semantic similarity focuses more on detailed information like the word "bakery" and the synonyms "muffins", "pastries", "cupcakes", and so on. (Bottom)The exemplar emphasizes the structural similarity between the exemplar and the question, which likely leads to a correct response.}
  \label{fig:demos}
\end{figure*}

An effective attempt at this problem is retrieval-based exemplar acquisition, where a retriever selects suitable exemplars based on input questions \cite{liu2021, unifyretrieve}. The retriever leverages encoders to encode examples and questions to vector representations, which enables a close exemplar selection. For example, \citeauthor{wu2022self} (\citeyear{wu2022self}) retrieve candidates with an off-the-shelf embedding model, followed by a re-ranking process. \citeauthor{epr} (\citeyear{epr}) consider the preference in the inner parameter space of language models. They evaluate that by the output probability of ground-truth label given question and demonstrations and fine-tuning the embedded with it. \citeauthor{ceil} (\citeyear{ceil}) consider both relevance and diversity for selection and instantiates that criteria with the Determinantal Point Process (DPP; \citeauthor{dpp}, \citeyear{dpp}). \citeauthor{dqlore} (\citeyear{dqlore}) notice the redundant information within the vector representation of exemplars and suggests a dimensionality reduction method Principal Component Analysis (PCA; \citeauthor{pca}, \citeyear{pca}) to tackle it.

Generally, these methods mainly concentrate on the semantic similarity between queries and examples in the training set. However, this metric measures superficial similarity by embedding the entire sentence into a specific vector space, which tends to ignore the hierarchical structure and can sometimes mislead the reasoning paths by focusing on local spurious relationships. The logical connection between intermediate steps on the reasoning path provides another type of important information \cite{graph-verify}. The logical connection can be abstracted to a graph structure, enabling a hierarchical representation that facilitates the calculation of similarity. These exemplars, characterized by structural information, are expected to guide language models along correct reasoning trajectories. For example, the top of Figure \ref{fig:demos} shows one case in which semantic similarity might partially focus on detailed and spurious relations, which leads to an error response. The bottom one in Figure \ref{fig:demos} demonstrates that a similar graph structure can induce LLM to generate a correct response. 

To take advantage of both semantic and structural information in the exemplar selection process, we propose a framework called Reasoning Graph Enhanced Exemplar Retrieval (RGER). RGER utilizes directed graphs to represent the reasoning pathways and, precisely, the topological relationships between intermediate steps. This method first retrieves candidate exemplars from a fine-tuned dense retriever, followed by re-ranking with a graph kernel metric for similarity calculation. After that, these selected instances are incorporated with semantic and structural similarity. These instances are composed of in-context learning. With the assistance of a graph structure, RGER can filter out redundant information and discern differences among various exemplars from a structural and hierarchical perspective. Extensive experiments are conducted on four math and two logic reasoning tasks. The result showcases that RGER can improve the performance of ICL compared to the latest retrieval-related methods. Furthermore, RGER exhibits robustness to tasks and language models' capability, showcasing its versatility across different scenarios. Our contributions can be summarized as follows:

\begin{itemize}
\item This paper introduces reasoning graphs for the exemplar selection process. The reasoning graph indicates the reasoning pathway and is critical to represent the problem-solving process.

\item This paper proposes Reasoning Graph-enhanced Exemplars Retrieval (RGER) for exemplar selection. RGER focuses on three aspects of the retrieval process, semantic relations, model preference, and structural relationship, which gives a thorough consideration in querying exemplars.

\item RGER demonstrates its strength to filter out redundant information from a structural perspective and achieve superior performance in different complex reasoning tasks, showcasing its efficacy and robustness across various scenarios.
\end{itemize}

\section{Related Work}

\begin{figure*}[!htbp]
  \centering
  \includegraphics[width=0.9\textwidth]{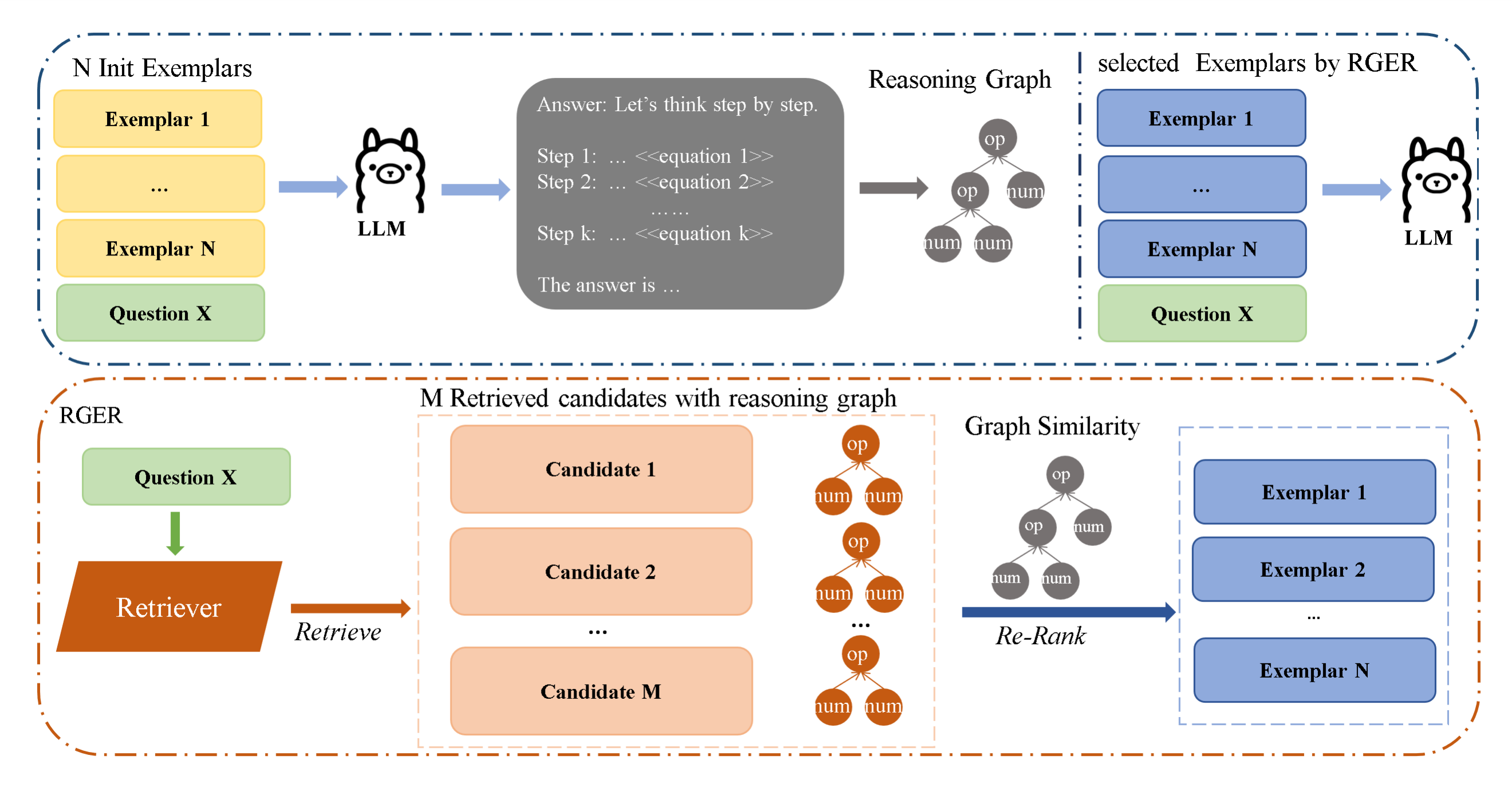}
  \caption{The pipeline overview of the proposed method. It consists of two parts: two-time queries (top) and retrieval process RGER (bottom).}
  \label{fig:framework}
\end{figure*}


\subsection{Exemplar selection for in-context learning}
Large language models (LLMs) empower the in-context learning (ICL) technique to solve different tasks with only a few demonstrations. Despite its effectiveness, studies have shown that performance can be easily influenced by the selected exemplars. A feasible solution for ICL is exemplar selection, where the retrieval-based paradigm is used to select the most appropriate demonstrations. Some works model the process with some off-the-shelf retrievers \cite{liu2021, unifyretrieve}. These methods retrieve exemplars with semantic similarity by sentence encoders. Other works consider the preference in the model’s representation space. For example, Efficient Prompt Retrieval (EPR; \citeauthor{epr}, \citeyear{epr}) uses the output probability given input and candidate demonstration to score the similarity between questions and exemplars. Based on that, EPR trains an efficient dense retriever. Compositional Exemplars for In-context Learning (CEIL; \citeauthor{ceil}, \citeyear{ceil}) instantiated by Determinantal Point Processes (DPPs;\citeauthor{dpp}, \citeyear{dpp}) to model the interaction between questions and exemplars considering both relevance and diversity. Dual Queries with Low-rank approximation Re-ranking (DQ-LoRe; \citeauthor{dqlore}, \citeyear{dqlore}) utilizes supplementary information from the generated text to subsequently re-query the retriever. Then re-rank candidates by dimensionality reduction to remove spurious
information within the embedding vectors.

\subsection{Graph Structure in Language Models}
Graphs encapsulate richly structured information and can represent a multitude of complex relationships, playing an essential role in improving the reasoning ability of LLMs. Some works enrich content representations by incorporating the structural information of graphs. For example, \citeauthor{graph-verify} (\citeyear{graph-verify}) trains a graph-augmented verifier to select correct responses generated by LLMs. \citeauthor{grace} (\citeyear{grace}) extracts graph structures from code data for software vulnerability detection. \citeauthor{graph-ensemble} (\citeyear{graph-ensemble}) propose a modal fusion pipeline with a graph encoder, which improves the reasoning ability of language models. 

Other works attempt to perform reasoning along the structural paths within graphs. \citeauthor{graph-guided} (\citeyear{graph-guided}) guides the model to generate knowledge triples along specific paths, leading to more reliable and accurate responses. Reasoning on Graph (RoG; \citeauthor{reason-on-graph}, \citeyear{reason-on-graph}) utilizes knowledge graphs to generate grounding paths as faithful paths. These paths are used to retrieve valid reasoning paths for faithful reasoning. Graph-of-Thought (GOT; \citeauthor{got}, \citeyear{got}) proposes a general framework to model the information generated by an LLM as arbitrary graph networks, which enables combining LLM thoughts into synergistic outcomes, distilling the essence of whole networks or enhancing thoughts using feedback loops.

\section{Methodology}

\begin{figure*}[!htbp]
  \includegraphics[width=0.95\linewidth]{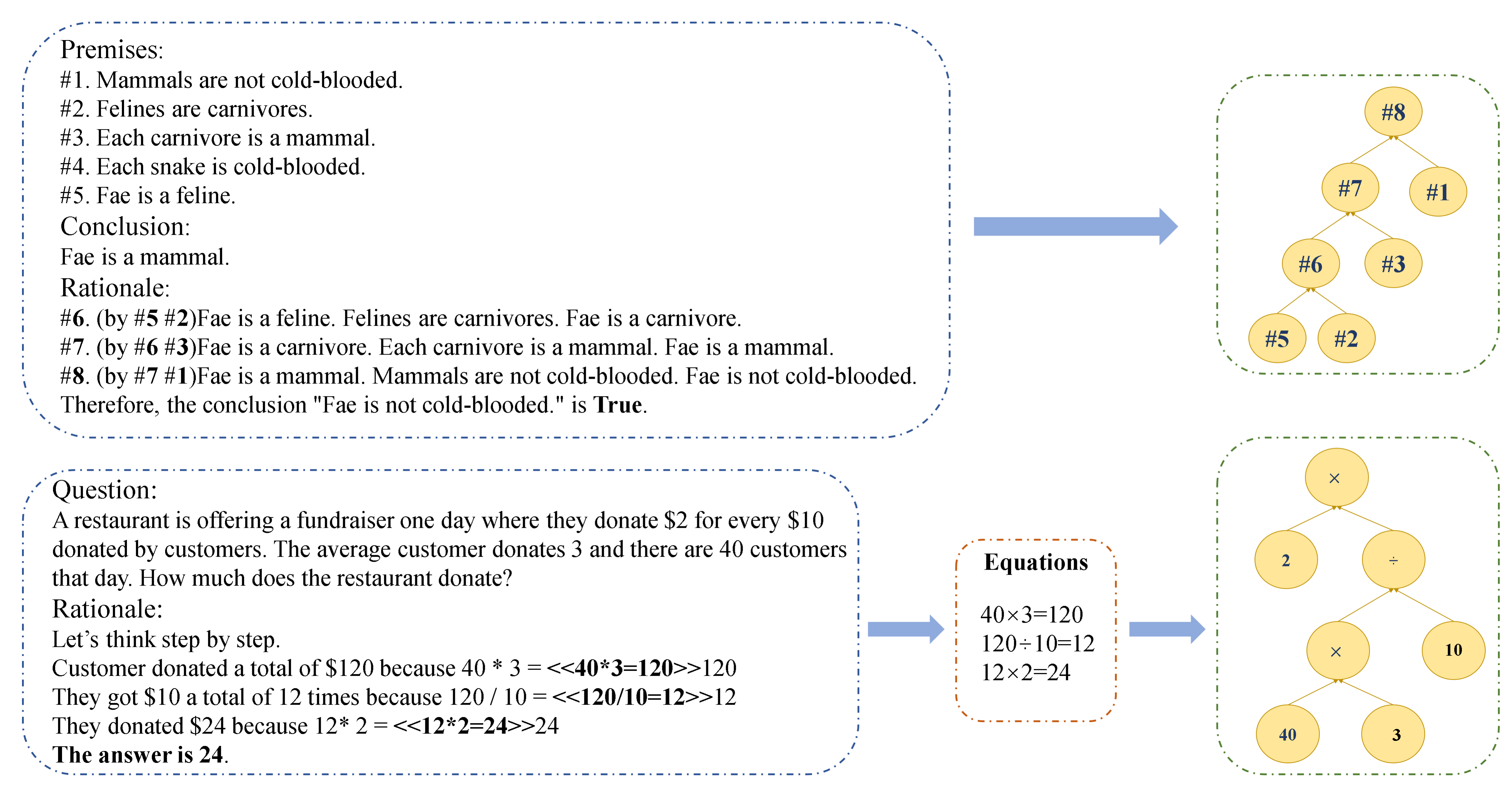}
  \caption {(Top) The approach prompts language models to generate a deductive reasoning form response, then extract the pathway to build a graph. (Bottom) The approach uses language models to extract equations from a response and then build a graph by Algorithm \ref{alg:algorithm}.}
  \label{fig:tograph}
\end{figure*}

\subsection{Reasoning with Structural Information}
As shown in Figure \ref{fig:framework}, our method consists of two parts. The first part contains two-time queries. We first query the LLMs to generate CoT with n initial exemplars. In this experiment, we simply employ the Complex-CoT approach \cite{complexcot} to obtain the initial demonstrations. This approach enables LLMs to engage in deeper reasoning and generate CoT with intermediate steps with richer information. Then, the responses are used to generate structural information represented as graphs. This information is used to retrieve the final exemplars. After the retrieval process, the selected exemplars are composed for the second time query.

The second part is the retrieval process. We designed a pipeline named Reasoning Graph Enhanced Exemplar Retrieval(RGER). RGER first queries the retriever by the question and chooses top-M candidates. Then, RGER utilizes the structural information for re-ranking and selects N exemplars for the final answers.

\subsection{Retriever Model Training}
This section introduces the way to train an encoder to represent exemplars and test samples that can measure the similarity between queries and exemplars. Similar to previous studies \cite{epr, ceil, dqlore}, contrastive learning is employed to train an encoder $E_p(\cdot)$ for similarity calculation. We utilize training set data, where each instance $d_i=(x_i,y_i)$ consists of a question $x_i$ and its corresponding answer with rationale $y_i$.

Following \cite{dqlore}'s setting, for each training sample $d_i$, the positive and negative set is constructed as follows: First, select top-k similar samples with an off-the-shelf retriever like BM25 \cite{bm25} and denoted as $D'=\{d_1',d_2',...,d_k'\}$. Then, $D'$ is re-ranked by considering how close the exemplar $d_j'$ is to the $d_i$ by the score function. The function is designed as the probability of LLM generating the CoT $y_i$ given the $d_j'$ and question $x_i$:
\begin{equation}
  score(d_j')=P_{LM}(y_i|d_j',x_i),\quad j=1,\dots,k  
\end{equation}

For sample $d_i$, select the top and last t samples as positive samples $pos_i$ and negative samples $neg_i$. For anchor point $d_i$ in a training batch, one positive sample $e_i^{+}$ is selected from $pos_i$, and one negative example $e_i^{-}$ is selected from $neg_i$ and examples of other anchors from the same batch. The contrastive loss with $b$ anchors can be expressed as follows:
\begin{equation}
    \begin{split}
      &Loss(x_i,x_j;e_i^+;e_1^+,e_1^-,...,e_i^-...,e_b^-) \\
      &=-\log\frac{e^{sim(x_i,x_j,e_i^+)}}{\sum_{j=1}^be^{sim(x_i,x_j,e_j^+)}+\sum_{j=1}^be^{sim(x_i,y_i,e_j^-)}}   
    \end{split}    
\end{equation}

where $sim(\cdot,\cdot,\cdot)$ is the inner product of the sequence embedding between anchor sample $d_i=(x_i,y_i)$ and an exemplar $d_j$.

\begin{algorithm}[tb]
\caption{Graph Structure Construction}
\label{alg:algorithm}
\begin{algorithmic}[1] 
\REQUIRE equation list $eqs = [eq_1,...,eq_k]$
\STATE Initialize G.
\FOR{$eq$ in $eqs$}
\STATE \textit{get answer $ans$ from $eq$.}
\STATE \textit{get an expression $exprs$ from $eq$.}
\STATE \textit{transfer $exprs$ to numbers, variables, and operands.}
\STATE \textit{Reorganize $exprs$ in a stack structure $rpn$ with Reverse Polish Notation.}
\WHILE{$rpn$ is not empty}
\STATE $lval$, $rval$, $op$ = $rpn.pop()$ 
\STATE $G.addnodes([lval,rval,op])$ 
\STATE $G.addedges((lval,op),(rval,op))$
\STATE $newval = Calcuate(lval,rval,op)$
\STATE $rpn.push(newval)$
\ENDWHILE
\STATE $G.node[op][value] = ans$
\ENDFOR
\STATE \textit{Check different properties of graph G.}
\STATE \textbf{return G}
\end{algorithmic}
\end{algorithm}

\subsection{Reasoning Graph Construction}
This section introduces two approaches to constructing reasoning graphs automatically. As shown in the left of Figure \ref{fig:tograph}, for reasoning tasks in general, the prompt for problem-solving can be designed in a deductive reasoning \cite{deductive} manner (shown in Appendix \ref{appendix:deductive} Listing ~\ref{deductive_prompt}). Then, by parsing the notations, the reasoning pathway from premises to intermediate conditions, and conclusions can be obtained effortlessly.

Another approach for abstracting math problem responses to graph structures is more complicated. As shown in the right of Figure \ref{fig:tograph}, with the aid of LLMs (shown in Appendix \ref{appendix:2eqs} Listing ~\ref{response2eqs}), calculation steps can be extracted and distilled into formulas from the original response. We found this conversion can be achieved by simple prompt demonstrations or a small-scale fine-tuned language model like opt-125m, which makes time and equipment costs negligible compared to LLM inference.

Algorithm \ref{alg:algorithm} summarizes an approach to construct a graph structure with a list of equations. For each equation, Algorithm \ref{alg:algorithm} draws edges from operands to result and saves operators to the result node iteratively. Reverse Polish Notation (RPN; \citeauthor{RPN}, \citeyear{RPN}) is introduced to cope with complex expressions like multiple operators or parentheses.  After that, the generated graph is scrutinized for different properties like strong or weak connectivity in graphs, connected components numbers, maximum path, and others. This can help rule out obvious errors in reasoning, such as repetition or inconsistency, which happen more often in smaller-scale generative language models.

\subsection{Graph Kernel as Re-ranking Metrics}
The graph kernel method provides an effective and straightforward method for measuring the similarity between graph structures \cite{metakernel}. It enables the capture of structural information within high-dimensional Hilbert spaces while maintaining the computational efficiency associated with kernel functions.

Most graph kernel methods are grounded in R-convolution theory \cite{metakernel}, which involves designing a decomposition method for graphs and evaluating the kernel between two graphs based on the similarity of their substructures. Given two graph $G_1(V_1,E_1)$ and $G_2(V_2,E_2)$, a graph decomposition method $\mathcal{F}$, the sub-structures are $\mathcal{F}(G_1)=\{S_{1,1},S_{1,2},...,S_{1,N_1}\}$ and $\mathcal{F}(G_2)=\{S_{2,1},S_{2,2},...,S_{2,N_2}\}$. Based on that, the kernel value can be expressed by 
\begin{equation}
k_R(G_1,G_2)=\sum_{n_1=1}^{N_1}\sum_{n_2=1}^{N_2}\delta(S_{1,n_1},S_{2,n_2}).
\end{equation}
Two notable kernel methods are the Shortest-Path Kernel (SP Kernel; \citeauthor{spkernel}, \citeyear{spkernel}) and the Weisfeiler-Lehman Kernel (WL Kernel; \citeauthor{wlkernel}, \citeyear{wlkernel}). The SP Kernel decomposes graphs into shortest paths and compares pairs of shortest paths according to their lengths and endpoint labels. On the other hand, the WL Kernel utilizes a vertex relabeling scheme where each vertex's label is replaced by a multiset label that includes the original vertex label and the sorted labels of its neighbors. This multiset is subsequently condensed into a new, shorter label.

In this study, we leverage graph kernel metrics to assess the structural similarity between the generated rationale of a problem and candidate exemplars. For math reasoning tasks, we apply the WL Kernel due to its capability to handle rich node attribute information. Specifically, we define three types of node attributes: numbers, variables, and operands. Conversely, for logic reasoning tasks where the reasoning graph lacks explicit node and edge information, the SP Kernel is more appropriate as it primarily focuses on graph structure.

\begin{table*}[htbp]
\caption{The accuracy(\%) of different models in four math reasoning tasks. All methods select 8-shot exemplars except for CoT, which uses 4-shot manually annotated exemplars.}
\centering
\begin{tabular}{clcccc|c}
\hline
Model                           & Method       & GSM8K          & AQUA           & SVAMP          & ASDIV          & Average        \\ \hline
\multirow{7}{*}{Llama2-7B-chat} & Manual-CoT\cite{cot}          & 23.05          & 19.29          & 58.33          & 53.13          & 38.45          \\
                                & Complex-CoT\cite{complexcot}  & \textbf{27.90} & 24.80          & 54.67          & 50.32          & 39.42          \\
                                & Auto-CoT\cite{autocot}     & 24.64          & 20.08          & 54.67          & 56.37          & 38.94          \\
                                & EPR\cite{epr}          & 23.12          & 22.44          & 57.00          & 56.59          & 39.79          \\
                                & CEIL\cite{ceil}         & 25.17          & 22.83          & 60.00          & 56.59          & 40.64          \\
                                & DQ-LoRe\cite{dqlore}      & 25.55          & 25.20          & 54.67          & 47.52          & 37.48          \\
                                & RGER(Ours)    & 26.23          & \textbf{25.59} & \textbf{61.33} & \textbf{60.26} & \textbf{43.35} \\ \hline
\multirow{7}{*}{Vicuna-7B}      & Manual-CoT\cite{cot}          & 19.26          & 18.50          & \textbf{55.00}  & 50.11          & 35.72          \\
                                & Complex-CoT\cite{complexcot}  & \textbf{22.67} & 24.02          & 43.00          & 46.36          & 34.01          \\
                                & Auto-CoT\cite{autocot}     & 20.62          & 24.41          & 50.00          & 55.08          & 37.53          \\
                                & EPR\cite{epr}          & 19.94          & 22.05          & 47.67          & 54.00          & 35.92          \\
                                & CEIL\cite{ceil}         & 18.42          & 25.20          & 49.00          & \textbf{55.29} & 36.98          \\
                                & DQ-LoRe\cite{dqlore}      & 22.29          & 22.44          & 47.67          & 49.24          & 35.41          \\
                                & RGER(Ours)    & 21.30          & \textbf{27.56} & \textbf{55.00} & 53.99          & \textbf{39.46} \\ \hline
\end{tabular}
\label{tab:mathres}
\end{table*}

\begin{table*}[htbp]
\centering
\caption{The accuracy(\%) in logic reasoning tasks ProntoQA and FOLIO with different methods. (Number in \textbf{bold} and \underline{underline} refer to the top and second highest numbers.)}
\begin{tabular}{lccccccc}
\hline
Dataset  & Manual-CoT & Complex-CoT & Auto-CoT & EPR           & CEIL & DQ-LoRe & RGER(Ours)    \\ \hline
ProntoQA & 89.6       &    93.6     & 92     & \textbf{94.6} & \underline{94.4} & 92.4    & \underline{94.4}          \\
FOLIO    & 57.1       &   45.3     & 61.1     & \underline{62.1}          & 61.6 & 43.3    & \textbf{62.6} \\ \hline
\end{tabular}
\label{tab:logic}
\end{table*}

\section{Experiments}
In the section, we introduce the experimental setup in Section \ref{setup}. We evaluate the proposed RGER method on both math reasoning and logic reasoning tasks in Section \ref{mainres}. We also consider a collaboration between LLMs by retrieving and inferencing in two distinct models in Section \ref{distribution_shift}. Further analysis in response quality(Section \ref{quality}) and ablation study(Section \ref{ablation}) are conducted as well. 

\subsection{Experimental Setup}\label{setup}
\subsubsection{Datasets}
We conduct comprehensive experiments on 4 math reasoning problem benchmarks, including GSM8K \cite{gsm8k}, AQUA \cite{aqua}, SVAMP \cite{svamp}, ASDIV \cite{asdiv} and 2 logic reasoning problem benchmarks ProntoQA \cite{prontoqa} and FOLIO \cite{folio}. Among that, ASDIV and FOLIO datasets lack reasoning rationales. Therefore we use GPT-4o\footnote{https://openai.com/index/hello-gpt-4o/} to generate responses in a chain-of-thought manner for retrieval. For ProntoQA, we follow the same setting as \citeauthor{rap} and generate 1500 instances varying from 3 to 5 hots. We also try our method in commonsence reasoning task StrategyQA \cite{strategyqa} in Appendix \ref{commonsense}.

\subsubsection{Evaluation}
Accuracy is used as the main metric to evaluate the performance of all tasks. For classification task AQUA, we select the option with the highest generation probabilities. For generation tasks, we set the temperature to 0 and employ greedy decoding for reproducibility. After that, the responses are applied with parsing rules in generated responses shown in Appendix \ref{sec:appendix_parserule}.

\subsubsection{Implementation}
\label{implementation}
For math reasoning tasks, our experiments are conducted on 3 LLMs: 2 open-source models, namely Llama2-7B-chat and Vicuna-7B, and 1 commercial model GPT-3.5-turbo-0125. For logic reasoning tasks, the experiments are tested on LLAMA-3-8B-Instruct.

We compare our method with 6 baseline methods, including Manual-CoT \cite{cot}, 2 heuristic exemplar selection methods: Complex-CoT \cite{complexcot}, Auto-CoT \cite{autocot}, and 3 latest retrieval-based methods: EPR \cite{epr}, CEIL \cite{ceil}, DQ-LoRe \cite{dqlore}. Appendix \ref{appendix:baseline} shows a detailed introduction of these baselines.

We employ demonstration examples and process code as suggested in the original papers for all baselines. In math reasoning tasks, the exemplar number is set to 8, except for Manual-CoT, which uses 4-shot exemplars. In the logic reasoning tasks, 4-shot is set as default, considering the context length limits. For graph similarity calculation of RGER, we utilize the WeisfeilerLehman kernel method \cite{wlkernel} in math reasoning tasks and the Shortest-Path kernel \cite{spkernel} for logical reasoning tasks. Implementation details can be found in Appendix \ref{sec:appendix_implement}.

\subsection{Main Result}\label{mainres}
Table \ref{tab:mathres} illustrates the performance of various models across four mathematical reasoning tasks, utilizing two backends: Llama2-7B-chat and Vicuna-7B. The results demonstrate that our proposed method achieves the highest average performance, with improvements of 6.7\% and 5.1\% over the baselines, respectively. Specifically, in the AQUA and SVAMP datasets, our proposed RGER method outperforms all other approaches. Additionally, in the GSM8K and ASDIV datasets, our method consistently exhibits a high level of performance, particularly when benchmarked against other retrieval-based techniques.

Table \ref{tab:logic} presents the performance of different methods across two logic reasoning tasks. In the relatively simpler ProntoQA task, RGER outperforms all manually designed methods and is competitive with other retrieval-based approaches. For the more complex FOLIO task, which requires longer reasoning steps, RGER also surpasses other baselines under the same experimental settings. Notably, compared to the approach DQ-LoRe, which directly concatenates questions and responses, RGER demonstrates superior information integration by extracting key topological information and representing it as graphs.



\begin{table}[t]
\centering
\caption{LLMs' Collaboration setting in closed-source LLM. The accuracy(\%) of different methods in SVAMP and GSM8K datasets. (Number in \textbf{bold} and \underline{underline} refer to the top and second highest numbers.)}
\label{tab:mathgpt}
\begin{tabular}{lcc}
\hline
Method        & SVAMP & \multicolumn{1}{l}{GSM8K} \\ \hline
ZeroShot-CoT  &  78.00    & 69.75                     \\
Manual-CoT    &  77.33    & 76.95                     \\
Auto-CoT      &  79.33    & 77.1                      \\
Complex-CoT   &  82.67    & 79.15                     \\
EPR           &  \textbf{86.67}    & 77.56                     \\
CEIL          &  82.33    & 77.71                     \\
DQ-LoRe       &  85.67    & 79.38                     \\
DQ-RGER(Ours) &  \underline{86.00}    & \underline{79.76}                     \\
RGER(Ours)    &  \textbf{86.67}    & \textbf{80.44}            \\ \hline
\end{tabular}
\end{table}

\subsection{Exemplar Selection for Closed-Source Model}\label{distribution_shift}

Commercial LLMs typically exhibit stronger capabilities in various areas \cite{gpt4}. However, their black-box property hinders access to model internal messages. In this section, we consider an applicable setting to improve the performance of closed-source LLM with exemplar selection methods by utilizing smaller open-source models(Llama3-8B-Instruct) to retrieve exemplars for large commercial models(GPT-3.5-Turbo-0125). This experiment explores the generalizability of different exemplar selection methods and provides insights into the collaboration between large-scale and smaller-scale LLMs. The result is shown in Table \ref{tab:mathgpt}.

This result shows the accuracy of SVAMP and GSM8K in different methods. We can see that retrieval-based methods achieve higher accuracy than other manual methods in general, which highlights the importance of exemplar selection. Our proposed method, RGER, outperforms all competing methods, showcasing its effectiveness and versatility. Besides, We designed a variation method DQ-RGER based on RGER, which utilizes the question and the corresponding first-time response as a query, just as DQ-LoRe does. DQ-RGER reaches the second-highest accuracy, which further showcases the generalizability of our reasoning graph-based selection method.

\subsection{Ablation Study}\label{ablation}

\begin{table}[t]
\centering
\caption{Ablation study for three main components in RGER: retriever, query field, and re-rank metric.}
\begin{tabular}{lcc}
\hline
\multirow{2}{*}{Method} & \multicolumn{2}{c}{GSM8K}       \\
                        & M=32           & M=64           \\ \hline
SenSim+q                 & 22.88          & 22.88          \\
SenSim+q+$\mathcal{G}$   & 25.83          & 25.55          \\
Retriever+q (EPR)      & 23.12          & 23.12          \\
Retriever+qa+$\mathcal{L}$ (DQ-LoRe)               & 24.94          & 22.52          \\
Retriever+qa+$\mathcal{G}$ (DQ-RGER)           & \textbf{25.85} & 24.64          \\
Retriever+q+$\mathcal{G}$ (RGER)               & 24.94          & \textbf{26.23} \\ \hline
\end{tabular}
\label{tab:ablation}
\end{table}

In this section, we provide a detailed analysis of the impact of each component of our method. Our method mainly consists of three components described as follows:
\begin{itemize}
    \item Retriever. The experiment compares the fine-tuned retriever and an off-the-shelf sentence-level embedding model all-MiniLM-L6-v2 \cite{sb}. In the experiment, we denote them as Retriever and SenSim, respectively.
    \item Query field for retrieve. We discussed the question-only type (q) and question and generated answers in a concatenated type (qa).
    \item Re-rank metric. Here, we compared two types of metrics: Low dimension similarity ($\mathcal{L}$, adhered to \citeauthor{dqlore}) and graph kernel similarity ($\mathcal{G}$). No re-rank is used as a control.
\end{itemize}

The experiment is conducted on GSM8K for accuracy(\%) with candidate numbers before re-ranking $M=\{32,64\}$ and Llama2-7b-chat as the backend.

The ablation experiment results are shown in Table \ref{tab:ablation}. RGER reaches high accuracy suggesting its efficacy among other compared methods. Three main components of RGER, trained retriever, query with question-only, and graph kernel for re-ranking, have average enhancements of 0.46, 1.59, and 2.12 to their counterparts. This result showcases the importance of each component. Besides, in $M=32$ settings, DQ-RGER performs well, which implies extra information in responses can be helpful to the retrieval process (But this is not stable as discussed in section \ref{mainres}). Furthermore, the result suggests the effectiveness and versatility of employing graph similarity to other methods. Appendix \ref{rger_theory} discusses the potential theoretical assurance for the method.

\begin{figure}[t]
  \centering
  \includegraphics[width=0.47\textwidth]{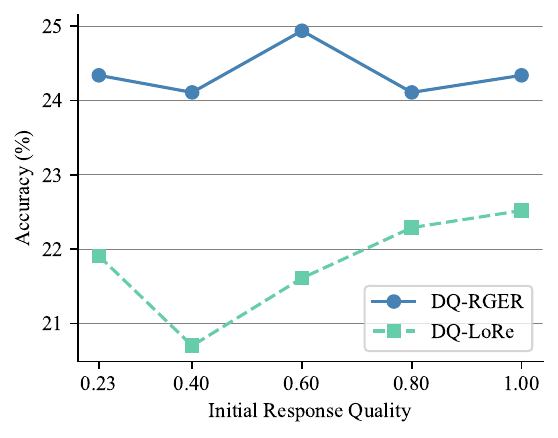}
  \caption{Comparison of two re-ranking methods, LoRe and Graph Similarity, under varying response quality levels.}
  \label{fig:dqcompare}
\end{figure}

\subsection{Influence in Response Quality}\label{quality}
Consider both our method and DQ-LoRe (\citeauthor{dqlore}) require an initial response for follow-up operations, emphasizing the importance of response quality. We experimented with those two approaches to investigate the relationship between initial response quality and the final response accuracy. We first mixture the responses from LLM-generated text and ground-truth answers to compose documents with different accuracy. These documents are utilized as criteria for retrieving exemplars. We concatenate the question and response as the retrieval query to make an intuitive comparison. To distinguish it from the original RGER method, we name it DQ-RGER. The experiment is conducted with $M=64$ under Vicuna-7b backend.

The result is shown in Figure \ref{fig:dqcompare}. We can see that the final accuracy for RGER is consistently higher than LoRe regarding re-ranking metrics. Compared to DQ-LoRe, DQ-RGER exhibits less fluctuation across different initial response qualities. With the rise of response quality, DQ-LoRe exhibits obvious improvement, while DQ-RGER appears more stable. The discovery further confirms our belief in designed pipeline in RGER: a). The reasoning graph has a relatively robust representation capability, and minor modifications to the initial response do not cause any degradation; b). To prevent a drastic impact on the final performance, our method begins by retrieving relevant exemplars to narrow the scope and then applies graph similarity for re-ranking.

\section{Conclusion}
This work introduces a novel exemplar selection approach tailored for complex reasoning tasks within the in-context learning (ICL) paradigm. We term this method Reasoning Graph Enhanced Exemplar Retrieval (RGER), which leverages topological relationships within reasoning steps for exemplar selection. By representing these relationships as graph structures and employing graph kernel methods to calculate similarity, RGER explicitly mitigates spurious correlations between problems and candidate exemplars. This approach facilitates a detailed and hierarchical selection process, moving beyond the simplistic embedding of entire sentences into semantic vector spaces. RGER guides the model to generate appropriate responses by following the reasoning path, thereby enhancing the model's capability to solve complex reasoning problems. Extensive experimental results demonstrate that RGER surpasses existing retrieval-based methods, showcasing superior efficacy, especially in tasks demanding intricate reasoning abilities.

\section{Limitations}

\textbf{Prompting and Fine-tuning.} RGER established an exemplar selection strategy for better performance in LLMs. Nonetheless, it is worthwhile to explore future endeavors of fine-tuning LLMs with this kind of high-quality data in a retrieval manner.

\noindent\textbf{Reasoning Graph Acquisition.} The reasoning graph is implicitly embedded within the text, and extracting it explicitly requires certain engineering methods, such as designing regular expression rules or prompting language models. We note that this is not a difficult task and can be achieved through an automated pipeline. However, there are still some outliers that do not align with our intentions. As to achieve optimal performance, human intervention is needed.

\noindent\textbf{Retriever Training.} Although our discoveries and previous work in \citeauthor{dqlore} find that commonalities existed in retrievers trained from relative tasks, it is more accurate to use the same LLM for inference as well as for the model-preference training. Unfortunately, there is some limitation to the generation probability of GPT-3.5-turbo yet\footnote{\url{https://platform.openai.com/docs/api-reference}}, while the optional completion models (such as text-davinci-003) are suspended. Considering the budget and machine limitations, we followed the principle and conducted our experiments with open-source models.

\section{Ethics Statement}
Large language models may occasionally generate biased or untrustworthy statements. Even though our intention is to improve the model, these issues are not being fully corrected. It is recommended that users exercise caution and refrain from overly relying on the model's outputs. This method is provided for research purposes only.

\section*{Acknowledgments}
This work is jointly supported by grants from 
the National Key R\&D Program of China (No. 2022ZD0116002),
Shenzhen Science and Technology Program (No. KCXFZ20230731093001002 and KJZD20230923115113026),
the Science and Technology Department of Guizhou Province (No. Qiankehe Support[2022]General019),
the National Social Science Foundation - Major Project (No. 20\&ZD226),
the Shenzhen Development and Reform Commission (No. XMHT20190108009),
the National Natural Science Foundation of China (No. 62276075 and 62102118),
the Guangdong Provincial Key Laboratory (No. 2022B1212010005),
Shenzhen Colleges and Universities Stable Support Program (No. GXWD20220811170504001 and GXWD20231129115148001),
Shenzhen Science and Technology Program (No. JCYJ20230807094318038),
Guangdong Basic and Applied Basic Research Foundation (No. 2023A1515110561),
the Key Laboratory of Intelligent Computing in Network Environment.


\bibliography{custom}

\clearpage
\appendix
\onecolumn
\section{Implement Details}\label{sec:appendix_implement}
We employ demonstration examples as suggested in the original papers for Manual-CoT and Complex-CoT. ZeroShot-COT uses the magic prompt "let's think step by step" \cite{zscot}. For tasks that were not mentioned in the original papers, we randomly chose demonstrations for Manual-CoT and selected examples with long reasoning steps for Complex-CoT from the training set. For Auto-CoT, we follow the algorithms from the original paper to select demonstrations from the training set. The candidate number before re-ranking M is set to 64 by default. All methods in math reasoning tasks select 8-shot exemplars except for Manual-CoT, which uses 4-shot exemplars. In logical reasoning problems, 4-shot demonstrations are employed considering the context length limits. 

For graph similarity calculation of RGER, we utilize the Weisfeiler-Lehman kernel method(Shervashidze et al., 2011) with 3 iteration steps in math reasoning tasks and Shortest-Path kernel(Borgwardt and Kriegel, 2005) for logical reasoning tasks.

In the retriever training process, we use Adam optimizer \cite{adam} with batch size 16, learning rate 1e-6 linear scheduling with 10\% warm-up. We train the model for 250 epochs on 1 NVIDIA A100-80G. For CEIL, we set re-ranking candidate number M to 100. For other retrieval-based methods, we fix the candidate number before re-ranking M to 64. For DQ-LoRe, we fix the dimension after PCA $\epsilon$ to 256.

Manual-CoT follows the prompt template in LINC \cite{linc}. Deductive-CoT follows the prompt template in \citeauthor{deductive} (\citeyear{deductive}) with the same instances in Manual-CoT. Deductive-CoT requires language models to generate responses with a deductive reasoning format.

 To make a fair comparison environment, we mainly focus on the exemplar selection strategies among different methods by setting the default retriever training rather than pursuing the highest performance.

\section{Parsing Rules}\label{sec:appendix_parserule}

\begin{table}[htbp]
\label{table:parserule}
\centering
\caption{Parsing rules format regular expression format, except for classification task AQUA, where we use perplexity of each answer choice.}
\begin{tabular}{ll}
\hline
Dataset  & Parsing Pattern                                                                  \\ \hline
GSM8K    & (-?\textbackslash{}d+(\textbackslash{}.\textbackslash{}d+)?)\textbackslash{}D*\$ \\
AQUA     & PPL of provided generation choices                                               \\
SVAMP    & (-?\textbackslash{}d+(\textbackslash{}.\textbackslash{}d+)?)\textbackslash{}D*\$ \\
ASDIV    & (-?\textbackslash{}d+(\textbackslash{}.\textbackslash{}d+)?)\textbackslash{}D*\$ \\
FOLIO    & (true|false|uncertain)                                                           \\ 
PrOntoQA & (true|false)  \\   \hline                                                               
\end{tabular}
\end{table}

\section{Experiment Result on Commonsense Reasoning Task}
\label{commonsense}
We further implemented our method in a commonsense reasoning task StrategyQA \cite{strategyqa}, which requires a multi-hot knowledge utilization ability for language models. We design a prompt shown in Listing ~\ref{strategyqa_prompt}  to extract structural and sentence-level information. Then we employ a simple heuristic graph re-ranking method for our RGER. The method can be described as follows:
$$Sim(q,e)=\alpha\cdot GraphSim(q,e)+(1-\alpha)\cdot Sensim(q,e)$$

$$Sensim(q,e)=SP\_Kernel(DAG(q),DAG(e))$$

$$Sensim(q,e)=\sum_{i=1}^m\max_{1\le j\le n}<q_i,e_j>=\sum_{i=1}^m\max_{j=1}^n\left(\begin{bmatrix}
    <q_1,e_1> & \ldots & <q_1,e_n> \\
    \vdots & \ddots & \vdots \\
    <q_m,e_1> & \ldots & <q_m,e_n>
\end{bmatrix}\right)$$

where $q:=\{q_1,...,q_m\}$ is the composite query instance with initial response and $e:=\{e_1,...,e_n\}$ is a candidate exemplar with steps. $SP\_Kernel$ refers to \citeauthor{spkernel} (\citeyear{spkernel}) discussed in Section \ref{implementation}. $Sensim$ is designed to allow the exemplar to semantically match the query at sentence level. $\alpha$ is the weight ratio, and we set it to 0.3 in this case.  

The result is shown in Table \ref{strategyqa}. Our method RGER outperforms all baseline methods.

\begin{table}[htbp]
\centering
\caption{Experiment result in StrategyQA for Llama2-7b-chat in 4-shot setting.}
\label{strategyqa}
\begin{tabular}{cc}
\hline
Method      & StrategyQA     \\ \hline
Manual-CoT  & 59.84          \\
Auto-CoT    & 61.14          \\
Complex-CoT & 64.63          \\
EPR         & 69.87          \\
CEIL        & 66.81          \\
DQ-LoRe     & 69.87          \\
RGER(Ours)  & \textbf{70.31} \\ \hline
\end{tabular}
\end{table}

\section{Potential theoretical assurance of RGER}
\label{rger_theory}
There are mainly two theoretical aspects to analyse our method RGER: \hyperref[theory_trans]{transformer perspective} and \hyperref[theory_info]{information theory}. 
\subsection{Information Theory Perspective}
\label{theory_info}
\citeauthor{wu2022self} (\citeyear{wu2022self}) formulate the exemplar selection process to a compression perspective and Minimum Description Length (MDL) from information theory. The process can be formulated as 

$$c^*=\arg\min_{c\in C}L_\theta(y|c,x)+L(\theta)$$
where each $c$ represents one possible organization of examples. $L_\theta(y|c,x)$ is the codelength required to compress and transmit testing label $y$ given the organization $c$ and testing input $x$. $L(\theta)$ is the codelength required to describe the model, which can be ignored during ranking. 

The codelength required for data transmission can be calculated using Shannon-Huffman code with an exact expression $L_\theta(y|c,x)=-\log_2p(y|c,x)$ and a proximate expression $L_\theta(y|c,x)\approx-\mathbf{E}_{q(y_i|Y)}\log_2(y_i|c,x)$. The final object is:
$$c^*=\arg\min_{c\in C}-\mathbf{E}_{q(y_i|c,x)}\log_2(y_i|c,x)$$
 
As discussed in \citeauthor{wu2022self} (\citeyear{wu2022self}), when we use the model confidence $p(y_i|c,x)$ as the estimation of $p(y_i|Y)$, the equation is basically calculating the entropy. Minimizing entropy is equivalent to searching for in-context examples that will lead to a skewed probability distribution. 

Our method proposes a fine-grained alignment method in hierarchical level, which makes consistency between query and exemplars or exemplars themselves. Thus, the composition text can approximate to the Minimum Description Length.

\subsection{Transformer Perspective}
\label{theory_trans}
Transformers have complex modules and multi-layer structures, making it challenging for the academic community to gain a comprehensive understanding of the mechanisms behind in-context learning. Recent work \cite{theoretical} has theoretically demonstrated that transformer is more sensitive to errors in intermediate steps than the final outputs, which stress the importance of exemplar selection process. RGER makes a great alignment between query and exemplars by integrating structural and semantic information in the selection process. It gives consistent intermediate solution steps for different tasks.

\section{Introduction of Our Baselines}\label{appendix:baseline}
We compare RGER to an extensive set of baselines and state-of-the-art methods, details are provided as follows:

\noindent\textbf{CoT} \cite{cot} has explored enhancing the learning process by manually designing a series of intermediate reasoning steps with in-context examples. This approach is applicable across various domains, such as mathematics, logic, symbols, and otherareasa requiring complex reasoning.

\noindent\textbf{Complex-CoT} \cite{complexcot} improves upon CoT for complex reasoning tasks by selecting exemplars with the most complex chain of thoughts. It achieves this by sampling multiple reasoning chains from the model and prioritizing the majority of generated answers from complex chains over simpler ones. Complex-CoT significantly boosts the accuracy of multi-step reasoning in LLMs reasoning performance.

\noindent\textbf{Auto-CoT} \cite{autocot} leverages the diversity within the training dataset to identify CoT with the greatest distinctions. The underlying principle is to extract a broader spectrum of information. Experimental results support this approach, demonstrating its effectiveness.

\noindent\textbf{EPR} \cite{epr} aims to retrieve prompts for in-context learning using annotated data with language models' preference. Given an input-output pair, EPR calculates the probability of the output given the input and a candidate training example as the prompt. Based on that, an efficient dense retriever is trained using the train data labeled as positive or negative samples by scores. After that, the retriever is employed to fetch exemplars as prompt demonstrations during testing. 

\noindent\textbf{CEIL} \cite{ceil} formulates the task of in-context example selection as a subset selection problem. It utilizes a determinantal point process(DPP) to capture the interaction between the given input and in-context examples. Through a carefully designed contrastive learning objective, CEIL manages to balance both the relevance of exemplars to the test questions and the diversity among the exemplars.

\noindent\textbf{DQ-LoRe} \cite{dqlore} design a dual-query framework that utilizes first-time response as extra knowledge from LLMs. A low-rank approximation re-ranking is employed to remove redundant or spurious relations implicitly. This method enhances the model’s ability to discern distinctions among various exemplars.

\section{More Examples for Building Reasoning Graph}
\begin{figure}[htbp]
  \includegraphics[width=\linewidth]{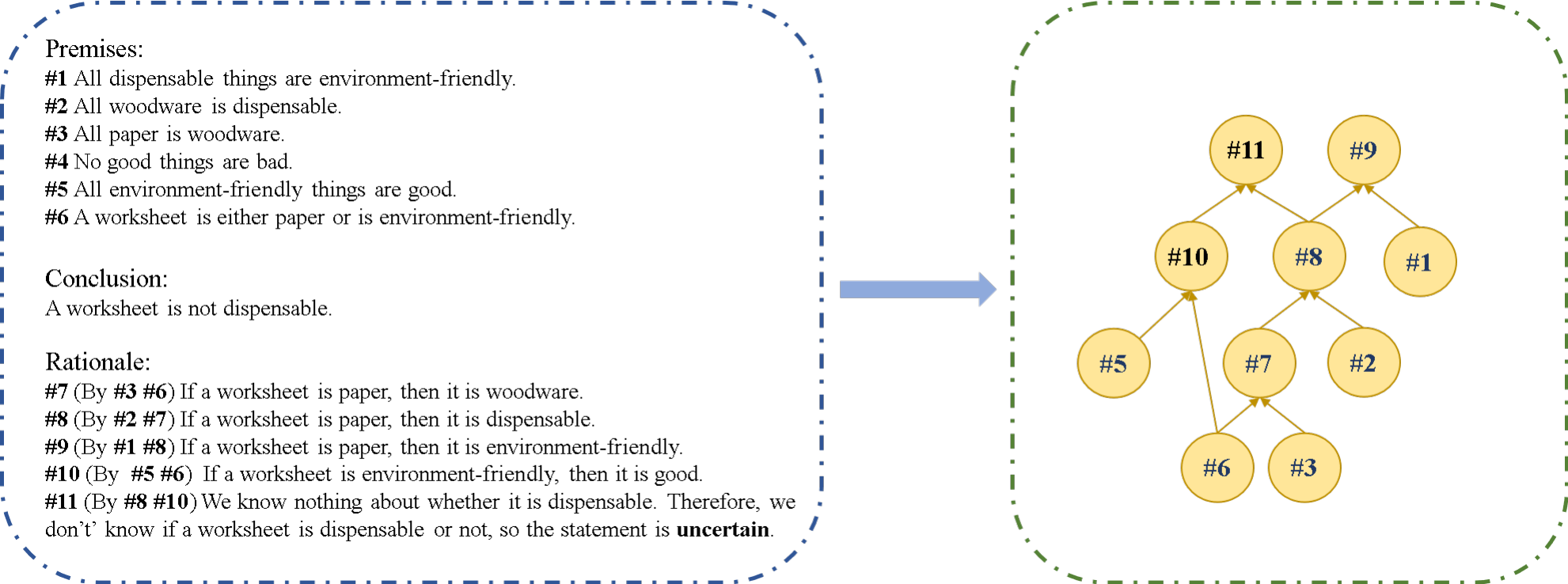}
  \caption {The approach prompts language models to generate a deductive reasoning form response, then extract the pathway to build a graph. }
  \label{appendix:logic}
\end{figure}

\begin{figure}[htbp]
  \includegraphics[width=\linewidth]{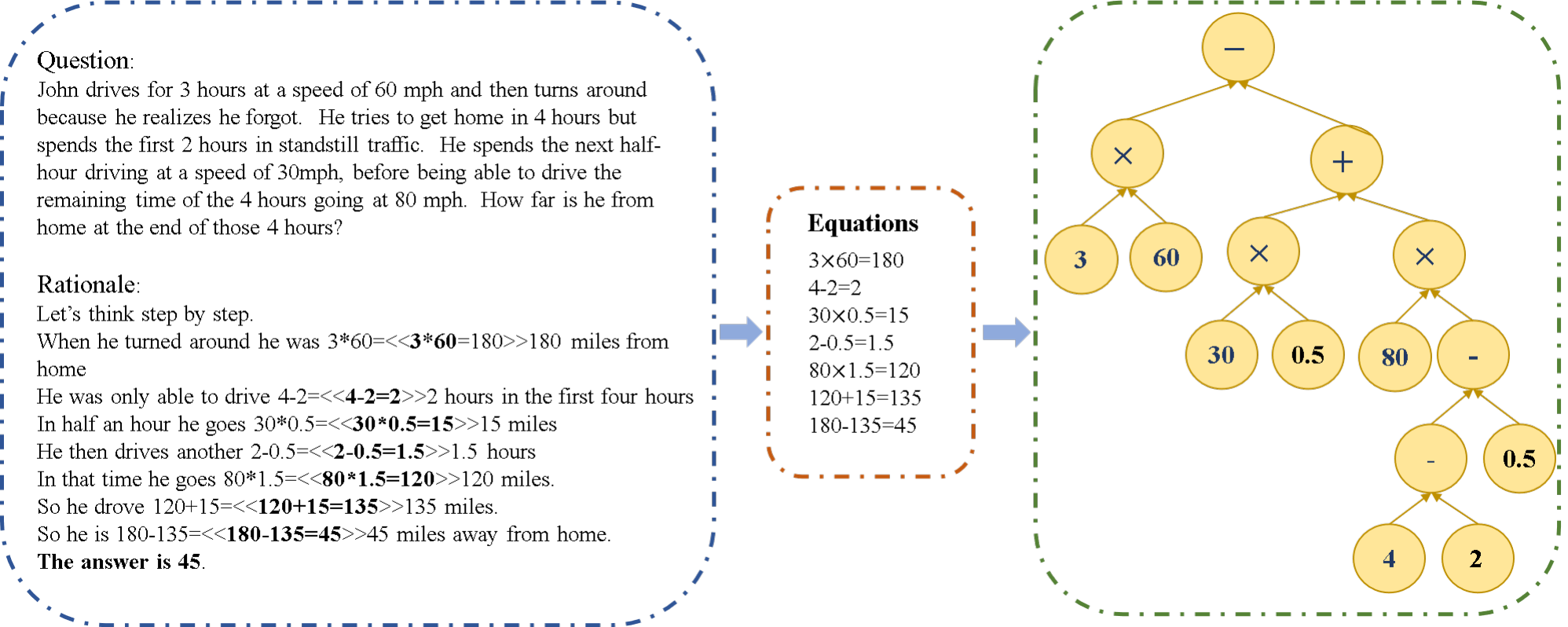}
  \caption {The approach uses language models to extract equations from a response and then build a graph by Algorithm \ref{alg:algorithm}.}
  \label{appendix:math}
\end{figure}

Figure \ref{appendix:logic} and figure \ref{appendix:math} demonstrate two more complex texts for the reasoning graph construction pipelines. The general expressive capability can be applied to different scenarios regardless of the complexity of the steps.

\section{Prompt Design}\label{appendix:prompt}

\definecolor{ao(english)}{rgb}{0.0, 0.5, 0.0}
\lstset{
frame=lines,
  aboveskip=3mm,
  belowskip=3mm,
  showstringspaces=false,
  columns=flexible,
  basicstyle={\footnotesize\ttfamily},
  numbers=none,
  numberstyle=\tiny\color{gray},
  keywordstyle=\color{blue},
  commentstyle=\color{dkgreen},
  stringstyle=\color{mauve},
  breaklines=true,
  breakatwhitespace=true,
  tabsize=2,
  moredelim=**[is][\color{red}]{@r}{@r},
  moredelim=**[is][\color{ao(english)}]{@g}{@g},
}

\subsection{Deductive Prompt}\label{appendix:deductive}
Listing ~\ref{deductive_prompt} shows an example of deductive prompt design. Based on that deductive pathway, a graph depicting the reasoning path can be easily extracted.

\begin{lstlisting}[caption={Deductive Prompts in ProntoQA}, label={deductive_prompt}]
Question: The following is a first-order logic (FOL) problem.
The problem is to determine whether the conclusion follows from the premises.
Here are premises:
Every integer is a real number.
Each Mersenne prime is not composite.
Prime numbers are natural numbers.
Every Mersenne prime is a prime number.
Real numbers are not imaginary.
Real numbers are numbers.
Every natural number is positive.
Each prime number is prime.
Every natural number is an integer.
Complex numbers are imaginary.
31 is a Mersenne prime.
Based on these premises, verify whether the conclusion "31 is imaginary." is True or False.

A: First, let's write down all the statements and relationships in the question with labels.
@g#1@g. Every integer is a real number.
@g#2@g. Each Mersenne prime is not composite.
@g#3@g. Prime numbers are natural numbers.
@g#4@g. Every Mersenne prime is a prime number.
@g#5@g. Real numbers are not imaginary.
@g#6@g. Real numbers are numbers.
@g#7@g. Every natural number is positive.
@g#8@g. Each prime number is prime.
@g#9@g. Every natural number is an integer.
@g#10@g. Complex numbers are imaginary.
@g#11@g. 31 is a Mersenne prime.
Next, let's answer the question step by step with reference to the question and reasoning process: 
@g#12. (by #11 #4)@g31 is a Mersenne prime.Every Mersenne prime is a prime number.31 is a prime number.
@g#13. (by #12 #3)@g31 is a prime number.Prime numbers are natural numbers.31 is a natural number.
@g#14. (by #13 #9)@g31 is a natural number.Every natural number is an integer.31 is an integer.
@g#15. (by #15 #1)@g31 is an integer.Every integer is a real number.31 is a real number.
@g#16. (by #15 #5)@g31 is a real number.Real numbers are not imaginary.31 is not imaginary.
Therefore, the conclusion "31 is imaginary." is False.
\end{lstlisting}

\subsection{Convert to Equations}\label{appendix:2eqs}
The prompt in Listing ~\ref{response2eqs} shows the extraction process from a rationale to math equations with the aid of a generated language model. We discover this is a simple ability and an LLM on a smaller scale like fine-tuned opt-125m \cite{opt} can achieve the requirement.

\begin{lstlisting}[caption={Prompts of converting a rationale to equations}, label={response2eqs}]
Given a response to a math problem, please tell all the calculation equations in order.
<example 1>
Response: If there are 290 bananas and they are organized into 2 groups, then each group would have 290/2 = <<290/2=145>>145 bananas.
Therefore, each group of bananas would have 145 bananas.
The answer is 145.
@gCalcuation equations: 290/2=145@g

<example 2>
Response: To determine how many more crayons Paul lost than those he gave to his friends, we need to subtract the number of crayons given to friends from the number of crayons lost. 
Number of crayons given to friends: 90
Number of crayons lost: 412
Difference: 412 - 90 = 322
Paul lost 322 more crayons than those he gave to his friends.
The answer is 322.
@gCalcuation equations: 412-90=322@g

<example 3>
Response: Edward spent $6 to buy 2 books, so each book cost $6/2 = $3.
Now he has $12, so the cost of the two books was $3 * 2 = $6.
Therefore, each book cost $3.
The answer is $3.
@gCalcuation equations: 6/2=3,3*2=6@g

<example 3>
Response: To find out how much Jessie weighed before starting to jog, we need to add the weight she lost to her current weight. 
Let's denote her previous weight as X. 
According to the information given, X - 126 = 66. 
To find X, we can add 126 to both sides of the equation: 
X - 126 + 126 = 66 + 126. 
This simplifies to: 
X = 192. 
Therefore, Jessie weighed 192 kilograms before starting to jog.
The answer is 192.
@gCalcuation equations: X-126=66,X=192@g

Following the above examples, please provide the calculation equations for the following responses:
\end{lstlisting}

\clearpage

\subsection{Prompts for StrategyQA}
\label{appendix:prompt_strategyqa}
We present the meticulously designed prompt in Listing ~\ref{strategyqa_prompt} which helps divide the response into steps.
\begin{lstlisting}[caption={Prompt example for StrategyQA}, label={strategyqa_prompt}]
Question: During the pandemic, is door to door advertising considered inconsiderate?
A: Let's think step by step
First, we will break down the problem into the following sub-problems:
#1 What does door to door advertising involve a person to do?
#2 During the COVID-19 pandemic, what does the CDC advise people to do in terms of traveling?
#3 During the COVID-19 pandemic, what does the CDC advise people to do in terms of interaction with others?
#4 Does doing #1 go against #2 and #3?

We have the following facts:
Door to door advertising involves someone going to several homes in a residential area to make sales and leave informational packets.
During the COVID-19 pandemic, the CDC recommends that people limit their travel to essential needs only.
During the COVID-19 pandemic, citizens are advised to stay home and to limit their interaction with others.
During the COVID-19 pandemic, people are encouraged to remain six feet away from each other at all times.
The more people that someone interacts with, the higher the likelihood of them becoming a vector for the COVID-19 virus.
Therefore, the answer is yes.
\end{lstlisting}

\section{Case Study on GSM8K}
In this section, we present the retrieve results for a single data point in GSM8K of applying different retrieval methods: EPR (Table \ref{cs_epr}), CEIL (Table \ref{cs_ceil}), DQ-LoRe (Table \ref{cs_dqlore1} and Table \ref{cs_dqlore2}), and our proposed RGER (Table \ref{cs_rger}). We can see through the demonstrations with some discoveries. EPR selects exemplars with co-occurrence words or proximity semantics to the question. In some cases, the retrieved examples have significantly different reasoning processes compared to the questions. CEIL select exemplars seem to be more diverse which is due to the DPP process. DQ-LoRe employs dimensionality reduction to implicitly filter out redundant information. The select examples seem to be more complicated and irrelevant to the question. But some of them possess a similar reasoning path to the answer, which is a two-step multiplication operation: "The fight lasted 5*3=<<5*3=15>>15 minutes. He threw 25*15=<<25*15=375>>375 punches. The answer is 375." RGER selects similar exemplars in the reasoning path aspect to the original questions. It explicitly filters out irrelevant information and differentiates examples based on structural similarity. RGER selects exemplars with similar problem-solving strategies providing appropriate guidance for LLMs.

\begin{table}[h]
    \caption{RGER Case Study on GSM8K}
    \label{cs_rger}
    \hrule
    \vspace{2pt}
    \textbf{Problem}
\\Question: On average Joe throws 25 punches per minute.  A fight lasts 5 rounds of 3 minutes.  How many punches did he throw?

\textbf{RGER Prompt}

Question: Tony lifts weights as a form of exercise.  He can lift 90 pounds with one arm in the exercise known as "the curl."  In an exercise known as "the military press," he can lift over his head twice the weight that he can curl.  His favorite exercise is known as "the squat" and he can squat 5 times the weight that he can lift in the military press.  How much weight, in pounds, can Tony lift in the squat exercise?
\\A: Let's think step by step If Tony can curl 90 pounds, he can military press 2*90=<<90*2=180>>180 pounds. Therefore, Tony can squat 5*180=<<5*180=900>>900 pounds. The answer is 900.

Question: Mike decides to do more pull-ups to increase his strength for climbing.  He uses the greasing the groove technique where every time he goes into a certain room he does 2 pull-ups.  He decides to use his office.  He goes in there 5 times a day every day.  How many pull-ups does he do a week?
\\A: Let's think step by step He does 5*2=<<5*2=10>>10 pull-ups a day So he does 10*7=<<10*7=70>>70 pull-ups a week The answer is 70.

Question: Rocky boxed 190 fights in his career.  50 percent of his fights were knockouts, and 20 percent of the knockouts were in the first round.  How many knockouts did Rocky have in the first round?
\\A: Let's think step by step Knockouts:190(.50)=95 First Round:95(.2)=19 The answer is 19.

Question: Marcus takes a deck of standard playing cards and takes out all the face cards and the 8's. Mark picks a card at random and then replaces it 36 times. How many times should he expect to pick a card that's both red and has a number divisible by 3?
\\A: Let's think step by step Without the face cards and the 8's, there are 9 cards of each suit. Half these cards are red, and 3/9 have numbers divisible by 3, so the odds of drawing a card divisible by 3 are 1/3. Multiply the odds of drawing a red card by the odds of drawing a card divisible by 3 to find the odds of both things happening: 1/3 * 1/2 = 1/6 Finally, multiplying the probability of drawing a red/3-divisible card by the number of cards Mark draws to find how many he should expect to meet his criteria: 1/6 * 36 cards = <<1/6*36=6>>6 cards The answer is 6.

Question: A bag of pistachios has 80 pistachios in it.  95 percent have shells, and 75 percent of those have shells that are opened.  How many pistachios in the bag have shells and have an opened shell?
\\A: Let's think step by step Shells:80(.95)=76 Opened Shells:76(.75)=57 The answer is 57.

Question: There are 500 students in a local high school.  40 percent are juniors.  70 percent of juniors are involved in sports.  How many juniors are involved in sports?
\\A: Let's think step by step Juniors:500(.40)=200 Juniors in sports:200(.70)=140 students The answer is 140.

Question: John hits 70\% of his free throws.  For every foul he gets 2 shots.  He gets fouled 5 times a game.  How many free throws does he get if he plays in 80\% of the 20 games the team plays?
\\A: Let's think step by step John plays in 20*.8=<<20*.8=16>>16 games That means he gets fouled 16*5=<<16*5=80>>80 fouls So he gets 80*2=<<80*2=160>>160 foul shots So he makes 160*.7=<<160*.7=112>>112 free throws The answer is 112.

Question: 40\% of the mosquitos in Jack's area are infected with malaria. 20\% of the mosquitos are infected with Zika virus. Without a vaccine, the chances of getting infected with either virus after getting bitten by an infected mosquito are 50\%. Jack is taking an experimental malaria vaccine that reduces the chances of getting infected after getting bitten by 50\%. If Jack gets bitten by a random mosquito, what is the percentage chance he catches either Zika virus or malaria?
\\A: Let's think step by step First find Jack's chances of getting bitten by an infected mosquito: 40\% + 20\% = 60\% The chance of being infected by a mosquito with after being bitten: 60\% * 50\% = 30\% Then find the chances of getting ill after receiving the vaccine: 30\% * 50\% = 15\% The answer is 15.

 \vspace{3pt}
\hrule
\end{table}

\begin{table}[h]
    \caption{EPR Case Study on GSM8K}
    \label{cs_epr}
    \hrule
    \vspace{2pt}
    \textbf{Problem}
\\Question: On average Joe throws 25 punches per minute.  A fight lasts 5 rounds of 3 minutes.  How many punches did he throw?

\textbf{EPR Prompt}

Question: Rocky boxed 190 fights in his career.  50 percent of his fights were knockouts, and 20 percent of the knockouts were in the first round.  How many knockouts did Rocky have in the first round?
\\A: Let’s think step by step. Knockouts:190(.50)=95 First Round:95(.2)=19 The answer is 19.

Question: Jeff plays tennis for 2 hours.  He scores a point every 5 minutes.  He wins a match when he scores 8 points.  How many games did he win?
\\A: Let’s think step by step. He played for 2*60=<<2*60=120>>120 minutes So he scores 120/5=<<120/5=24>>24 points That means he won 24/8=<<24/8=3>>3 games The answer is 3.

Question: John manages to run 15 mph for his whole 5-mile race.  The next fastest guy ran the race in 23 minutes.  How many minutes did he win the race by?
\\A: Let’s think step by step. He runs the race in 5/15=1/3 of an hour So he runs the race in 60/3=<<60/3=20>>20 minutes That means he beat the next fastest guy by 23-20=<<23-20=3>>3 minutes The answer is 3.

Question: Tim is stuck in traffic for twice as long as he was driving.  He drove 5 hours.  How long was the trip?
\\A: Let’s think step by step. He was stuck in traffic for 5*2=<<5*2=10>>10 hours So his trip took 10+5=<<10+5=15>>15 hours The answer is 15.

Question: James drives to Canada at 60 mph.  It is a distance of 360 miles.  He has a 1 hour stop along the way.  How long does he take to get to Canada?
\\A: Let’s think step by step. He spends 360/60=<<360/60=6>>6 hours driving That means the trip takes 6+1=<<6+1=7>>7 hours The answer is 7.

Question: Bob buys nose spray.  He buys 10 of them for a "buy one get one free" promotion.  They each cost \$3.  How much does he pay?
\\A: Let’s think step by step. He got 10/2=<<10/2=5>>5 paid That means he paid 5*3=\$<<5*3=15>>15 The answer is 15.

Question: Matt skips ropes 3 times per second.  If he jumped for 10 minutes how many skips hops did he get?
\\A: Let’s think step by step. He was skipping rope for 10*60=<<10*60=600>>600 seconds So he got 600*3=<<600*3=1800>>1800 skips The answer is 1800.

Question: John plays at the arcade for 3 hours.  He uses \$.50 for every 6 minutes.  How much money did he spend, in dollars?
\\A: Let’s think step by step. He was at the arcade for 3*60=<<3*60=180>>180 minutes So he put coins in 180/6=<<180/6=30>>30 times That means he spent 30*.5=\$<<30*.5=15>>15 The answer is 15.

    \vspace{3pt}
    \hrule
\end{table}

\begin{table}[h]
    \caption{CEIL Case Study on GSM8K}
    \label{cs_ceil}
    \hrule
    \vspace{2pt}
    \textbf{Problem}
    \\Question: On average Joe throws 25 punches per minute.  A fight lasts 5 rounds of 3 minutes.  How many punches did he throw?

\textbf{CEIL Prompt}
Question: Lily types 15 words a minute and takes a 2-minute break every 10 minutes. How long does it take for Lily to type 255 words?
\\A: Let’s think step by step. It would take Lily 255 words / 15 wpm = <<255/15=17>>17 minutes to type without taking a break. Since Lily takes a break after 10 minutes of typing she takes 17 minutes + 2 minutes = 19 minutes The answer is 19.

Question: Rocky boxed 190 fights in his career.  50 percent of his fights were knockouts, and 20 percent of the knockouts were in the first round.  How many knockouts did Rocky have in the first round?
\\A: Let’s think step by step. Knockouts:190(.50)=95 First Round:95(.2)=19 The answer is 19.

Question: Jeff plays tennis for 2 hours.  He scores a point every 5 minutes.  He wins a match when he scores 8 points.  How many games did he win?
\\A: Let’s think step by step. He played for 2*60=<<2*60=120>>120 minutes So he scores 120/5=<<120/5=24>>24 points That means he won 24/8=<<24/8=3>>3 games The answer is 3.

Question: Paul goes fishing every Saturday. Last week he was able to catch 5 fish for every 2 hours he was fishing. How many fish did he catch when he was fishing for 12 hours?
\\A: Let’s think step by step. Paul caught 5 fish 12 / 2 = 6 times that day. So he was able to catch in total 6 * 5 = <<6*5=30>>30 fish. The answer is 30.

Question: George and Harry want to fill a pool with buckets. George can carry two buckets each round, and Harry can carry three buckets each round. If it takes 110 buckets to fill the pool, how many rounds will it take to fill the pool?
\\A: Let’s think step by step. Each round 2+3=5 buckets will be filled. It will take 110/5=<<110/5=22>>22 rounds. The answer is 22.

Question: John decides to get the vaccine for COVID.  He has to wait 20 minutes for the first dose.  The second dose has a wait time half as long.  How long was the total wait time?
\\A: Let’s think step by step. The second dose had a wait time of 20/2=<<20/2=10>>10 minutes So the total wait time is 20+10=<<20+10=30>>30 minutes The answer is 30.

Question: Pete walked 5 blocks from his house to the bus garage in Houston. He rode the bus 20 blocks to the post office to get some stamps. Later, he came home the same way. How many blocks did Pete travel in all?
\\A: Let’s think step by step. The number of blocks traveled on the way to the post office was 5 + 20 = <<5+20=25>>25. Pete traveled 25 × 2 = <<25*2=50>>50 blocks in all. The answer is 50.

Question: The Period 1 gym class has 5 fewer than twice as many students as in the Period 2 gym class. There are 11 students in the Period 1 gym class. How many are in the Period 2 gym class?
\\A: Let’s think step by step. Let x be the number of students in the Period 8 gym class. 2x-5=11 2x=16 x=<<8=8>>8 The answer is 8.

    \vspace{3pt}
    \hrule
\end{table}

\begin{table}[h]
    \caption{DQ-LoRe Case Study on GSM8K: Part 1}
    \label{cs_dqlore1}
    \hrule
    \vspace{2pt}
    \textbf{Problem}
\\Question: On average Joe throws 25 punches per minute.  A fight lasts 5 rounds of 3 minutes.  How many punches did he throw?

\textbf{DQ-LoRe Prompt}
Question: A fox can run at the maximum speed of 50 kilometers per hour. Considering the fox would run at a constant speed, what distance would he make during 120 minutes?
\\A: Let’s think step by step. 1 hour is 60 minutes, so 120 minutes is 120 / 60 = <<120/60=2>>2 hours. During one hour a fox can make 50 kilometers, so during two hours, he would be able to run 50 * 2 = <<50*2=100>>100 kilometers. The answer is 100.

Question: Jacob can write twice as fast as Nathan. Nathan wrote 25 letters in one hour. How many letters can the two write in 10 hours together?
\\A: Let’s think step by step. Since Nathan writes 25 letters in an hour, Jacob writes 25*2 = <<25*2=50>>50 letters in an hour. Together, they write 50+25 = <<50+25=75>>75 letters in an hour, In 10 hours, they'll write a total of 75*10 = <<10*75=750>>750 letters The answer is 750.

Question: To get to work, Ryan bikes one day a week takes the bus three times a week and gets a ride from a friend once a week. It takes him thirty minutes to bike to work. The bus takes ten minutes longer since it stops at every bus stop for other passengers. His friend driving him is the fastest and cuts two-thirds off his biking time. How many minutes does he spend every week commuting to work?
\\A: Let’s think step by step. The bus to Ryan’s work takes 10 minutes longer than his bike ride, so it takes 10 + 30 = <<10+30=40>>40 minutes. Ryan takes the bus 3 times a week, so the bus rides take 40 * 3 = <<40*3=120>>120 minutes a week. His friend cuts two-thirds off his biking time, which is 2/3*30 = <<2/3*30=20>>20 minutes a week. This means his friend takes 30-20 = <<30-20=10>>10 minutes to drive him. Therefore, Ryan spends 120 + 10 + 30 = <<120+10+30=160>>160 minutes every week commuting to work. The answer is 160.

Question: Five times a week, Onur bikes 250 kilometers a day. His friend Hanil bikes 40 more kilometers of Onur biking distance in a day. What's the total distance the two friends bikes in a week?
\\A: Let’s think step by step. Since Onur bikes 250 kilometers in a day, he bikes 5*250 = <<250*5=1250>>1250 kilometers for five days in a week. Onur's friend Hanil bikes 40 more kilometers in a day than Onur, meaning he bikes for 250+40 = <<250+40=290>>290 kilometers. In five days of the week, Hanil covers 290*5 = <<290*5=1450>>1450 kilometers. In total, the two bikes for 1450+1250 = <<1450+1250=2700>>2700 kilometers in 5 days of a week. The answer is 2700.

Question: Vivian is responsible for making sure her students get 2 15-minute recess breaks a day, a 30-minute lunch and another 20-minute recess break.  How much time do her students spend outside of class?
\\A: Let’s think step by step. They get 2 15 minute breaks a day so that’s 2*15 = <<2*15=30>>30 minutes They get 30 minutes for lunch and 20 minutes for recess plus an additional 30 minutes of recess for a total of 30+20+30 = <<30+20+30=80>>80 minutes spent outside of class The answer is 80.
    \vspace{3pt}
    \hrule
\end{table}

\begin{table}[h]
    \caption{DQ-LoRe Case Study on GSM8K: Part 2}
    \label{cs_dqlore2}
    \hrule
    \vspace{2pt}
    \textbf{Problem}
\\Question: On average Joe throws 25 punches per minute.  A fight lasts 5 rounds of 3 minutes.  How many punches did he throw?

\textbf{DQ-LoRe Prompt}

Question: Hank reads the newspaper every morning, 5 days a week for 30 minutes.  He reads part of a novel every evening, 5 days a week, for 1 hour.  He doubles his reading time on Saturday and Sundays.  How many minutes does Hank spend reading in 1 week?
\\A: Let’s think step by step. He reads 30 minutes in the morning 5 days a week for a total of 30*5 = <<30*5=150>>150 minutes He doubles his reading time on the weekends so 30*2 = <<30*2=60>>60 minutes per day He reads 60 minutes on Saturday and Sunday in the mornings for a total of 60+60 = <<60+60=120>>120 minutes He reads 1 hour in the evenings 5 days a week for a total of 1*5 = <<1*5=5>>5 hours He doubles his reading time on the weekends so 1*2 = <<1*2=2>>2 hours per day He reads 2 hours on Saturday and Sunday in the evenings for a total of 2+2 = <<2+2=4>>4 hours He read 5+4 = 9 hours which, when converted to minutes is 9*60 = 540 minutes He reads 150 minutes in the morning and 120 minutes in the evenings and 540 over the weekend for a total of 150+120+540 = 810 minutes in 1 week The answer is 810.

Question: Grayson has 100 questions on a test. He answers some of the questions for 2 hours, taking two minutes for each answer, then realizes he doesn't know the rest of the questions. How many questions does Grayson leave unanswered when he submits his test?
\\A: Let’s think step by step. Each hour contains 60 minutes, so after 2 hours he took 2*60 = <<2*60=120>>120 minutes. If each question takes two minutes, then he answered 120/2 = <<120/2=60>>60 questions. Unanswered questions is 100-60 = <<100-60=40>>40 questions The answer is 40.

Question: Micah can type 20 words per minute and Isaiah can type 40 words per minute. How many more words can Isaiah type than Micah in an hour?
\\A: Let’s think step by step. There are 60 minutes in an hour. Micah can type 20 x 60 = <<20*60=1200>>1200 words in an hour. Isaiah can type 40 x 60 = <<2400=2400>>2400 words in an hour. Isaiah can type 2400 - 1200 = <<2400-1200=1200>>1200 words more than Micah in an hour. The answer is 1200.
    \vspace{3pt}
    \hrule
\end{table}

\end{document}